\documentclass[letterpaper, 10 pt, conference]{ieeeconf} 
\IEEEoverridecommandlockouts                         
\usepackage{graphics}
\usepackage{amsmath}
\usepackage{amssymb} 
\usepackage{bm}
\usepackage{graphicx}
\usepackage{comment}
\usepackage[hidelinks]{hyperref}

\title{\LARGE \bf
Omnidirectional Dual-Arm Aerial Manipulator with Proprioceptive Contact Localization for Landing on Slanted Roofs
}

\author{
Martijn B.J. Brummelhuis$^{1}$, Nathan F. Lepora$^{2}$ and Salua Hamaza$^{1}$
\thanks{$^{1}$Martijn Brummelhuis and Salua Hamaza are with the BioMorphic Intelligence Lab, Dept. Control \& Operations, Faculty of Aerospace Engineering, TU Delft, The Netherlands. \newline
E-mail correspondance: {\tt\small m.b.j.brummelhuis-1@tudelft.nl}     
}%
\thanks{$^{2}$Nathan Lepora is with the School of Engineering Mathematics and Technology, and Bristol Robotics Laboratory, University of Bristol, United Kingdom}
\thanks{Video available at: \url{https://youtu.be/V3gcCDMcqBQ}}
}

\begin{document}

\maketitle
\thispagestyle{empty}
\pagestyle{empty}

\begin{abstract}
Operating drones in urban environments often means they need to land on rooftops, which can have different geometries and surface irregularities. Accurately detecting roof inclination using conventional sensing methods, such as vision-based or acoustic techniques, can be unreliable, as measurement quality is strongly influenced by external factors including weather conditions and surface materials. \\
\indent To overcome these challenges, we propose a novel unmanned aerial manipulator morphology featuring a dual-arm aerial manipulator with an omnidirectional 3D workspace and extended reach. Building on this design, we develop a proprioceptive contact detection and contact localization strategy based on a momentum-based torque observer. This enables the UAM to infer the inclination of slanted surfaces blindly -- through physical interaction -- prior to touchdown. We validate the approach in flight experiments, demonstrating robust landings on surfaces with inclinations of up to 30.5$^{\circ}$ and achieving an average surface inclination estimation error of 2.87$^{\circ}$ over 9 experiments at different incline angles.
\end{abstract}

\section{Introduction}
Unmanned Aerial Manipulators (UAMs) have advanced considerably in recent years, however most works focus heavily on single specific operations. There is a need for versatility so that the UAM can handle different situations that may occur during a mission such as perching, landing, and manipulation. This versatility can be achieved by using the limbs as a Swiss knife - a single tool with various different applications. \\
\indent Research into aerial manipulators has demonstrated many possible morphologies to expand on UAMs' versatility.
Most commonly, aerial manipulators employ serial robotic arms \cite{ollero2021past} or parallel mechanisms \cite{bianconi2025forcebalanced}; both of which are well suited for pick-and-place tasks. More recently continuum arms are making an appearance \cite{peng2025dexterous}. Dual-arm and multi-arm aerial manipulators \cite{suarez2018design, paul2021tams} extend these capabilities by enabling more complex manipulation, grasping of long or voluminous objects \cite{caballero2018first}, or even using one arm as an anchor point while the other performs a task. \\
\indent Many aerial manipulator designs have their limbs mounted rigidly on a single location on the body, limiting the robot's workspace and therefore versatility \cite{ryll20196d, staub2018towards}. To expand the reachable workspace and enable a broader range of applications, aerial vehicles with omnidirectional workspaces have been proposed. Most of these platforms are fully or over-actuated systems that rely on \textit{fixed-tilt} rotors \cite{park2018odar} or \textit{variable-tilt} rotors, allowing the thrust vector to be reoriented in the desired direction without incurring efficiency losses \cite{allenspach2020design}. While these designs achieve workspace omnidirectionality by exploiting the motion of the aerial platform itself to control a rigidly attached manipulator, they require significantly more complex control strategies to handle unconventional -- and, in the case of tilting rotors, time-varying -- thrust allocation. Few approaches address this limitation by decoupling manipulation from drone platform orientation, enabling the manipulator to operate within a workspace that revolves planarly 360$^{\circ}$ around the vehicle’s body \cite{hamaza2020omni, ollero2019aerox}.

\begin{figure}[t]
  \centering
  \includegraphics[width=\linewidth]{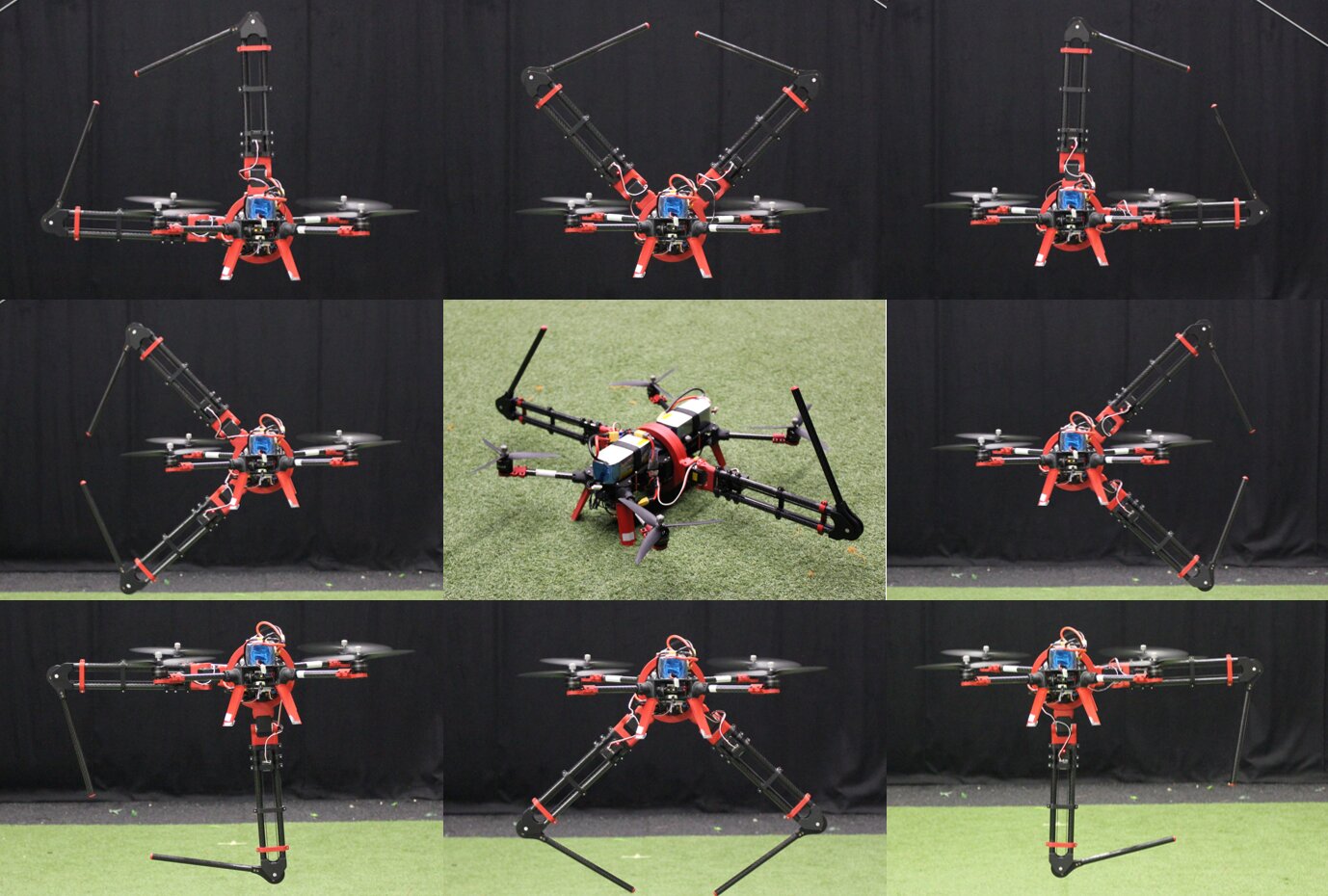}
  \caption{Mosaic image of the dual-arm aerial manipulator in different configurations, showcasing the omnidirectional workspace.}
  \label{fig:mosaic}
\end{figure}

\indent A scenario showing the need for versatility is landing in urban areas. Slanted roofs are especially common and offer a safe option, since they keep the drone away from pedestrian areas and traffic. The challenge is that drones with standard, unidirectional thrust, e.g. quadrotors, cannot easily align with the angle of an inclined roof. For larger (and heavier) aerial robots in particular, coming in at an angle and then "dropping" onto the surface \cite{bass2022adaptative} is risky, as the hard impact can damage both the drone and the roof. Other approaches based on advanced motion planning \cite{thomas2016aggressive} and deep learning \cite{kooi2021inclined} to align the drone with the surface at touchdown have been proposed at high computational cost, prohibiting on-board deployment. Another method to mitigate the hard landing by using a tethered approach has been shown \cite{tognon2016takeoff}, but this presumes a method for tethering close to the landing point, which is not a trivial condition to satisfy.\\
\indent Sensing methods such as cameras \cite{kim2021autonomous} or laser rangefinders \cite{tsintotas2021safe} can struggle in poor weather or low-light conditions, while acoustic sensors depend heavily on how well the roof reflects sound. The shortcomings of the aforementioned sensing methods can be avoided by using contact to infer the landing location geometry. Earlier work exploits rich force information to obtain explicit stiffness measurements of target branches \cite{aucone2023drone}, but does not allow inferring the environment's geometry. More generally, all these approaches fundamentally depend on exteroceptive sensing hardware, motivating solutions that do not rely on external perception to infer landing geometry.\\
\indent Even when the landing surface properties are successfully estimated -- whether through vision, range, or contact-based sensing, the drone may still be unable to perform the landing due to mechanical constraints posed by the uneven ground. Several works use adaptive landing gear to overcome this issue. A mechanically simple approach using hinged skids allows landing on inclines \cite{kim2021autonomous}. More complex landing surfaces can be handled with multiple landing legs \cite{liu2021multi, ramirez2023bioinspired, choi2021robust}. These mechanisms address landing only, whereas aerial manipulators require multifunctional limbs.\\
\indent To address these problems we present three contributions:
\begin{itemize}
    \item A novel dual-arm aerial manipulator design with 3D omnidirectional workspace and low-inertia limbs that minimize the reaction torques on the body;
    \item A proprioceptive, sensorless, contact detection and contact localization which enables the aerial manipulator to learn the inclination of landing slopes;
    \item Flight experiments validating the effectiveness of the omnidirectional morphology and proprioceptive pipeline for the landing on surfaces of unknown inclination.
\end{itemize}

\section{Modeling}

\subsection{Kinematics}
\begin{figure}
    \centering
    \includegraphics[width=0.95\linewidth]{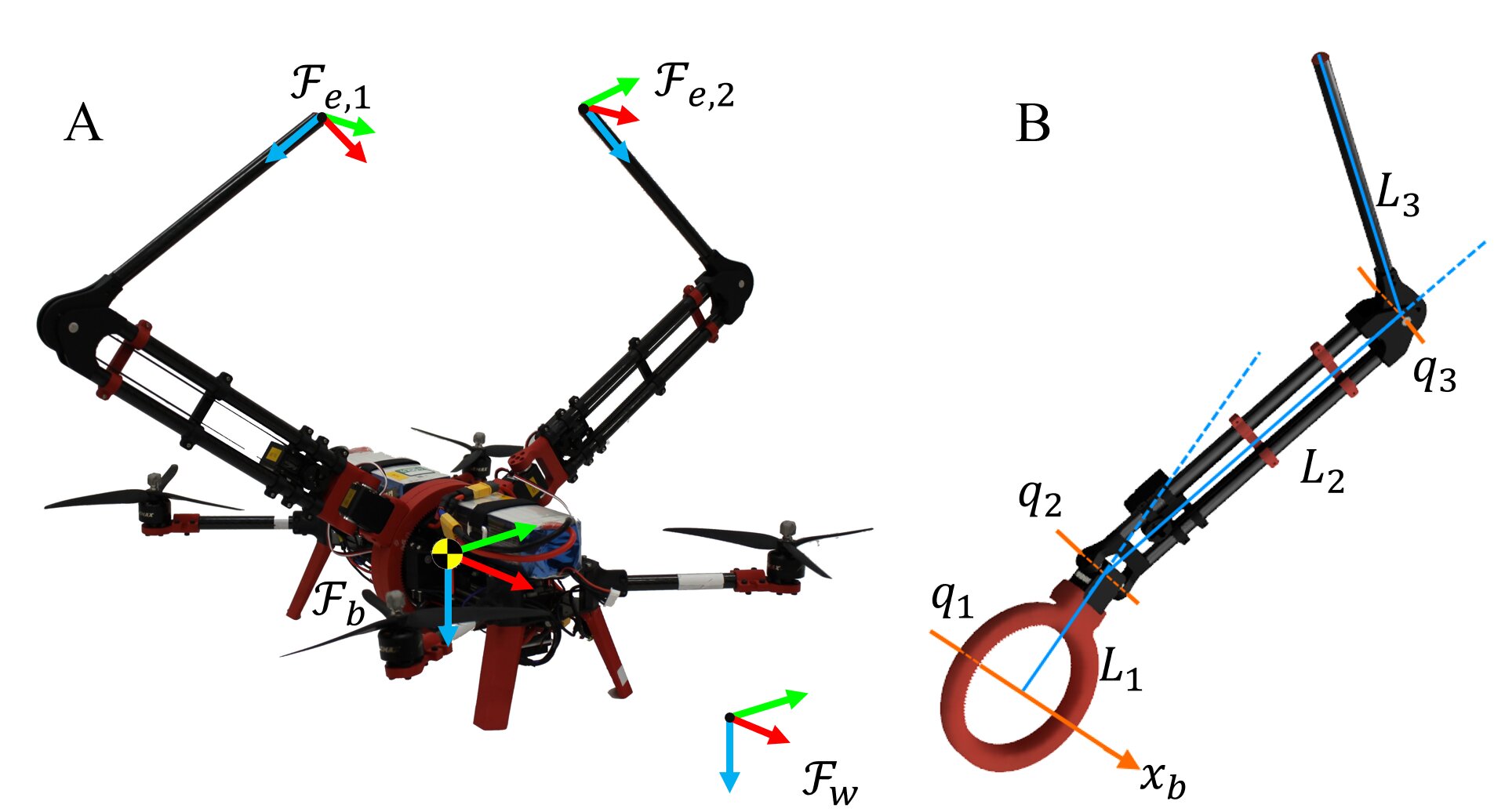}
    \caption{A. The dual-arm omnidirectional aerial manipulator and its main reference frames with axes colored red, green, and blue (RGB) for the X-, Y-, and Z-axes, respectively, B. Kinematics of a single manipulator arm, with axis of rotation of $q_1$ coinciding with the body x-axis.}
    \label{fig:frames}
\end{figure}

A North-East-Down (NED) world frame $\mathcal{F}_{w}$ is attached to the ground. The UAM has its body frame $\mathcal{F}_{b}$ attached at the CoM following the Forward-Right-Down (FRD) convention. The end-effector frames $\mathcal{F}_{e,j}$ with $j=1,2,...,k$ (up to $k$ arms) are attached at the end of the respective arms as shown in Figure \ref{fig:frames}.A. The generalized state of a multi-arm aerial manipulator $k$ arms of $i=1,2,...,n$ DoFs each is $\bm{\xi}=[\bm{p}_{b} \;\; \bm{\phi}_{b} \;\; {}^{1}\bm{q} \;... {}^{k}\bm{q}]^{\top}\in \mathbb{R}^{6+nk}$, with $\bm{p}_{b}=[x_{b} \; y_{b} \; z_{b}]^{\top}\in \mathbb{R}^{3}$ the position of $\mathbf{O}_{b}$ in $\mathcal{F}_{w}$, $\mathbb{\phi}_{b}=[\varphi_{b} \; \theta_{b} \; \psi_{b}]^{\top}$ the orientation of $\mathcal{F}_{b}$ in $\mathcal{F}_{w}$ represented by roll, pitch, and yaw angles, and ${}^{j}\bm{q}=[^{j}q_{1} \; ... \; ^{j}q_{n}]^{\top} \in \mathbb{R}^{n}$  the configuration vector of manipulator arm $j$ .\\
\indent The kinematics can be expressed in homogeneous transformation matrices. The pose of the body is denoted $H^{w}_{b}([\bm{p}_{b} \; \; \bm{\phi}_{b}]^{\top}) \in SE(3)$ and the pose of end-effector $j$ in the body frame is denoted $H^{b}_{e,j}({}^{j}\bm{q})$, which is a function of the joint positions ${}^{j}\bm{q}$ of serial arm $j$. 
The world pose of the end-effectors can then be obtained by pre-multiplication of their pose in the body frame with the pose of the body in the world
\begin{equation}
    H^{w}_{e,j}(\bm{\xi}) = H^{w}_{b}([\bm{p}_{b} \; \; \bm{\phi}_{b}]^{\top}) H^{b}_{e,j}({}^{j}\bm{q}).
\end{equation}
The expressions for the position forward kinematics in the body frame $\bm{p}^{b}_{e,j}$ can be extracted from the homogeneous transformation matrix $H^{b}_{e,j}$.\\
\indent The proposed aerial manipulator features two manipulator arms of 3 revolute joints. Figure \ref{fig:frames}.B showns one arm, along with the joint and link names. The axis of rotation of the first joint coincides with the body x-axis.

\subsection{Inverse Kinematics}
Since the manipulator has 3 DoFs, only the inverse kinematics for end-effector position and linear velocity control are considered here. The position inverse kinematics are derived algebraically.\\
\indent The differential kinematics are expressed using the Jacobian as function of the joint positions $J({}^{j}\bm{q})$:
\begin{equation}
    \dot{\bm{p}}^{b}_{e,j}=\bm{v}^{b}_{e,j}=J({}^{j}\bm{q}){}^{j}\dot{\bm{q}},
\end{equation}
with $\bm{v}^{b}_{e,j}$ the linear velocity vector in the body frame of end-effector $j$. The linear differential inverse kinematics can be obtained straightforwardly by inverting $J({}^{j}\bm{q})$ at a given manipulator configuration ${}^{j}\bm{q}$:
\begin{equation}
    {}^{j}\dot{\bm{q}}=J^{-1}({}^{j}\bm{q})\bm{v}^{b}_{e,j}.
\end{equation}
This is done by evaluating $J({}^{j}\bm{q}){}^{j}\dot{\bm{q}}$ at a certain ${}^{j}\bm{q}$ and then inverting, rather than calculating an algebraic inverted Jacobian matrix, for computational reasons.
\subsection{Dynamics}
Utilizing the conventions of the previous subsections, the dynamics of the aerial manipulator can be modeled \cite{lippiello2012cartesian} to obtain the equations of motion in the form
\begin{equation}
    M(\bm{\xi})\ddot{\bm{\xi}}+C(\dot{\bm{\xi}},\bm{\xi})\dot{\bm{\xi}}+G(\bm{\xi})=\bm{\tau}_{ext}+N\bm{u}.
\end{equation}
In this expression, the mass-inertia matrix is denoted $M(\bm{\xi})\in\mathbb{R}^{(6+nk)\times(6+nk)}$, the matrix of centrifugal and Coriolis forces is $C(\dot{\bm{\xi}},\bm{\xi})\in\mathbb{R}^{(6+nk)\times(6+nk)}$, and the gravity effects are denoted by $G(\bm{\xi})\in\mathbb{R}^{6+nk}$. The non-conservative forces are split into external forces $\bm{\tau}_{ext}$ and the actuator forces $N\bm{u}$, with $N$ the allocation matrix mapping the input vector $\bm{u}$ onto the generalized coordinates.

\section{Aerial Manipulator Design}
The dual-arm aerial manipulator has been designed for a large and omnidirectional workspace, while keeping the mass and inertia of the limbs low to reduce the effect of moving masses on the UAM body. \\
\indent The body consists of an upper and lower carbon plate, with blocks mounting the rotor arms on the outer corners, providing a stiff box structure with space for mounting electronics. Dual 6000 mAh batteries are mounted on the top side of the platform to balance the weight. The body has a standard quadrotor configuration and uses a Pixhawk 6X flight controller. The design thrust-to-weight ratio is 2, which is confirmed in the flight tests. The UAM's most important properties are detailed in Table \ref{tab:drone_specs}.

\subsection{Pivot joint}
\begin{figure}[t]
    \centering
    \includegraphics[width=\linewidth]{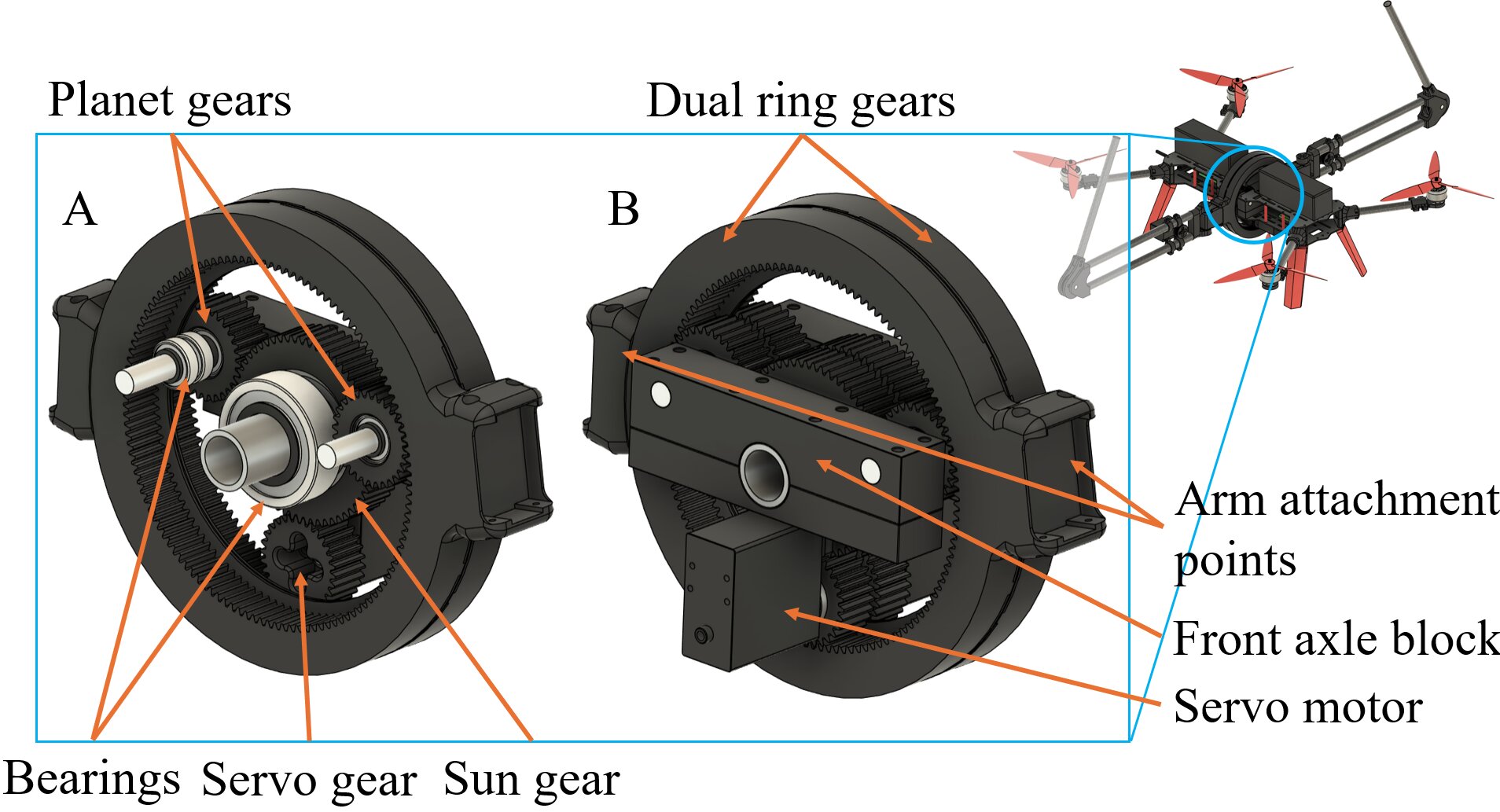}
    \caption{Close up view of the pivot joint mechanism that enables omnidirectional motion about the UAM's centre of mass. A) shows key components inside the front axle block, including the planetary gear mechanism. B) shows the full pivot joint assembly with the axle blocks and servo.}
    \label{fig:pivotjoint}
\end{figure}

The two arms are mounted on a pivot joint inspired by \cite{hamaza2020omni}, allowing both arms to rotate fully around the body and operate in any direction relative to the body.\\
\indent The mechanism resembles a planetary gear, with one of the planets being driven instead of the central sun gear. The sun and non-driving planetary gears allow axes to pass through and carry the weight of the manipulator arms, see Figure \ref{fig:pivotjoint}. \\

\begin{table}[b]
    \centering
    \caption{Properties of the presented UAM, with components present twice on the UAM indicated with 2$\times$.}
    \begin{tabular}{l|r || l|r}
    \hline
        \textbf{Property} & \textbf{Value} & \textbf{Property} & \textbf{Value}\\ \hline
        Body length $l_b$ & 0.352 m & Body mass & 1287 g \\
        Body width $w_b$ & 0.110 m & Battery mass & 2$\times$ 828 g \\
        Rotor arm length $l_r$ & 0.160 m & Link 1 mass &2$\times$ 236 g \\
        Link 1 length $L_1$ & 0.118 m & Link 2 mass & 2$\times$ 304 g\\
        Link 2 length $L_2$ & 0.330 m &  Link 3 mass & 2$\times\;\;$ 61 g  \\
        Link 3 length $L_3$ & 0.273 m & Total mass & 4145 g \\
        Propeller diameter & 9" (0.229 m) & Battery type & 2$\times$6S 6Ah \\
        Propeller type & 9045 & Motor type & 2814 830KV \\\hline
    \end{tabular}
    \label{tab:drone_specs}
\end{table}
\begin{figure}[t]
    \centering
    \includegraphics[width=\linewidth]{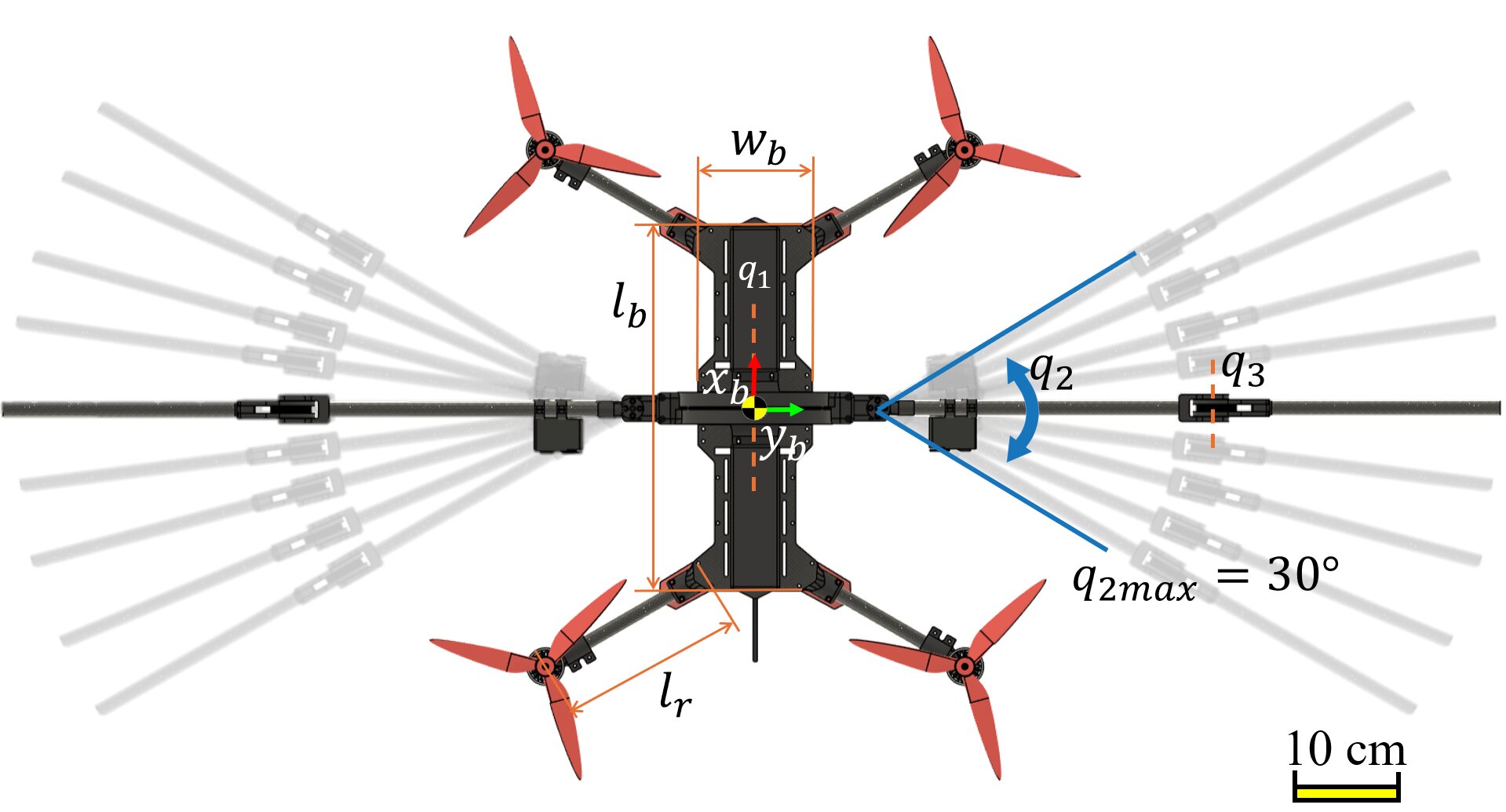}
    \caption{Top view of the aerial manipulator's out-of-the-plane motion. }
    \label{fig:out-of-plane-top}
\end{figure}
\subsection{Manipulator arms}
\begin{figure}
    \centering
    \includegraphics[width=\linewidth]{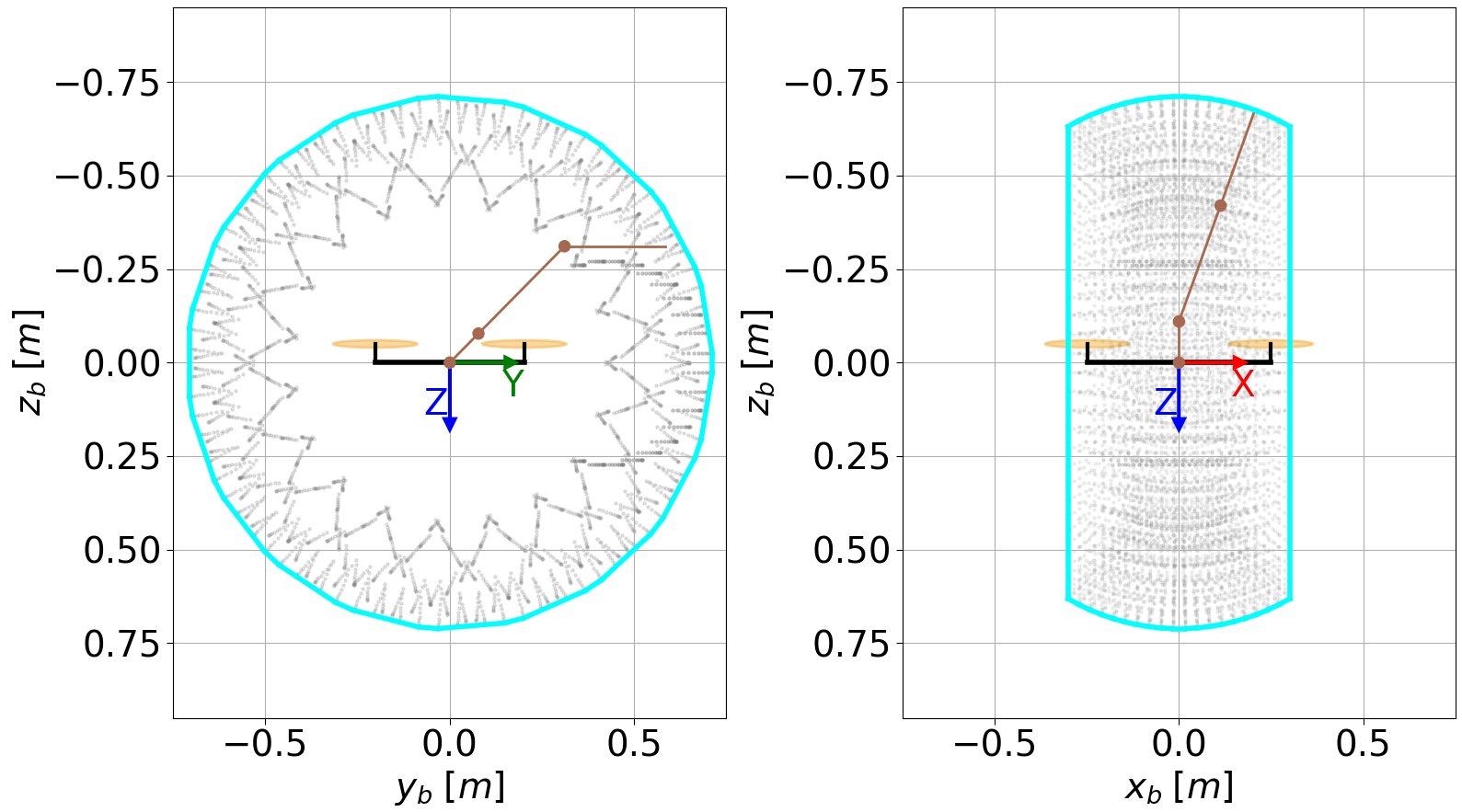}
    \caption{Three-dimensional workspace plots of the novel aerial manipulator. The blue line marks the workspace outline and the grey points are sampling points. The left plot shows the workspace in the YZ-plane. The right plot shows the workspace in the XZ-plane, i.e. the out-of-plane reach.} \label{fig:workspace}
\end{figure}

The manipulator arms are 3R serial arms with the second joint allowing a movement out of the rotation plane of the pivot joint, see Figure \ref{fig:out-of-plane-top}. Joints 1 (pivot) and 3 (elbow) have parallel axes of rotation when there is no out-of-plane motion from joint 2. The inertia of the arms about the vehicle's CoM is lowered by using drive belts to transfer torque from the servo to the outer joint. This allows for placing the servo closer to the vehicle's CoM, similar to \cite{katz2019mini}. Crosslinks in the upper arm provide structural stiffness and torsion resistance, allowing the lightweight manipulator to carry the UAM's weight when landing. The joints are actuated by smart servos (Feetech ST3025 , STS3250, and STS3235, from body to end-effector) providing position, velocity, and current feedback. \\
\indent A workspace analysis of the system (by sampling the configuration space within the joint limits) is presented in Figure \ref{fig:workspace}, showing the omnidirectional and large workspace of the aerial manipulator, along with a schematic of the aerial manipulator platform where the length, width, and rotor diameter are real relative size.
\section{Proprioceptive contact detection and localization}
\begin{figure*}
    \centering
    \includegraphics[width=\linewidth]{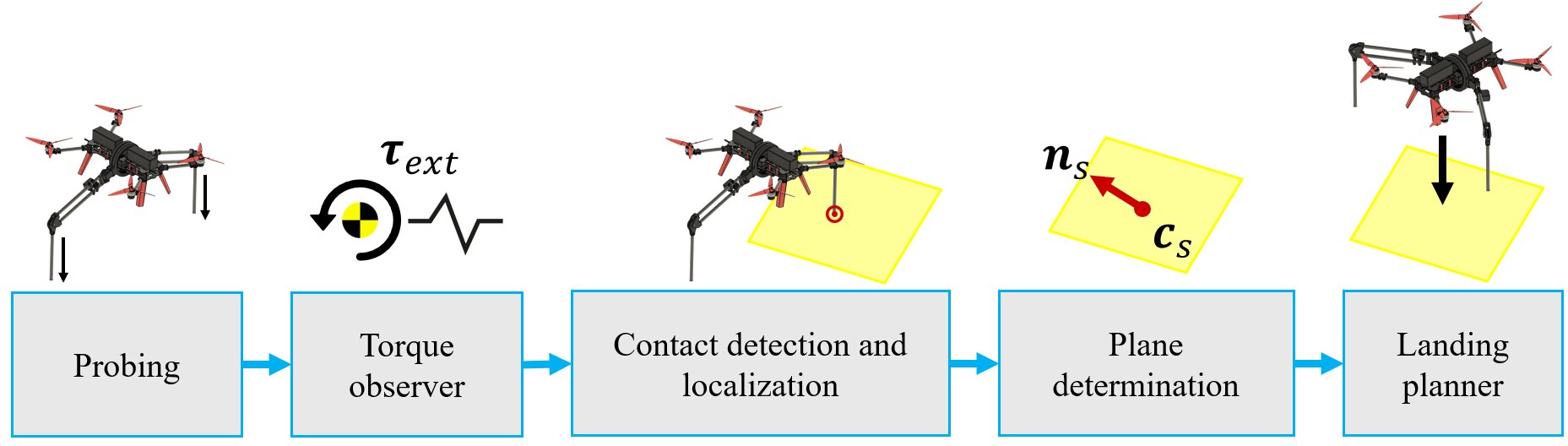}
    \caption{Flowchart showing the pipeline that enables safe landing on inclined surfaces.}    \label{fig:pipeline}
\end{figure*}
Given the low force/torque transparency of high-gear-ratio servo motors, determining which manipulator collides with an object through current feedback is difficult, especially for momentary impact loads and floating bodies. To detect when contact occurs and on which arm, external torques are estimated from angular velocity measurements coming directly from the gyro on the IMU by a momentum-based external torque observer.\\
\indent The total angular momentum of the system at a given time instant is
\begin{equation}
    \bm{L}(t) = I^{b}_{b}\bm{\omega}^{b}_{b}(t) + \sum^{}_{i=0}\left( I^{i}_{i}\bm{\omega}^{i}_{i}(t) + m_{i}\bm{v}^{b}_{i}(t)\right),
\end{equation}
where $I^{b}_{b}$ denotes the mass moment of inertia of the body about the system CoM, $\bm{\omega}^{b}_{b}$ the angular velocity of the body in the body frame, $I^{i}_{i}$ the mass moment of inertia of link $i$ about its own CoM, $m_{i}$ the mass of link $i$, and $\bm{v}^{b}_{i}$ the linear velocity of link $i$ in the body frame. The expression is simplified by considering that the masses and moments of inertia of the arm are small compared to the body's and that in quasi-static conditions the velocities also remain small. This allows the simplification of only considering the body for the contact detection. Inertia parameters are obtained from an accurate Computer-Aided Design (CAD) model which was verified by checking that part masses from a weighing scale correspond to part masses in the model. \\
\indent The torque is then estimated as the residual of the change of angular momentum at any given time instant $t$ \cite{ryll20196d}:
\begin{equation}
\begin{aligned}
    \hat{\bm{\tau}}^{b}_{b}(t) = & \bm{L}(t)-\bm{L}(t_{0}) + \\
    & \int^{t}_{t_{0}}(\bm{\omega}^{b}_{b}(\tau)\times I\bm{\omega}^{b}_{b}(\tau)-N_{O}\bm{u}(\tau)-\hat{\bm{\tau}}^{b}_{b}(\tau))d\tau,
\end{aligned}
\end{equation}
where $\hat{\bm{\tau}}^{b}_{b}$ is the external torque estimate and $N_{O}$ denotes the part of the allocation matrix corresponding to the body moments. This expression can then be discretized for implementation as
\begin{equation}
\begin{aligned}
    \hat{\bm{\tau}}^{b}_{b}(t_{d})= & I\bm{\omega}^{b}_{b}(t_{d}) + \\
    & \left(\bm{\omega}^{b}_{b}(t_{d})\times I\bm{\omega}^{b}_{b}(t_{d})-N_{O}\bm{u}(t_{d})\right)\delta t + \\
    & \sum^{t_{d}-1}_{i=0}\left( \bm{\omega}^{b}_{b}(i)\times I \bm{\omega}^{b}_{b}(i)-N_{O}\bm{u}(i)- \hat{\bm{\tau}}^{b}_{b}(i)  \right)\delta t,
\end{aligned}
\end{equation}
by realizing that the initial angular momentum can be set to zero as the UAM is at rest at $t=0$, discretizing the integral and taking out the last term of the summation. The discretized time at present is denoted $t_{d}$ with $\delta t$ being the time per step in seconds.\\
\indent A contact point is registered if the absolute value of the estimated torque rises above a threshold and a timeout prevents registration of multiple contact points due to bouncing. Contact points are only registered when the UAM is looking for contact to discard false positives.\\
\indent For contact localization, a virtual moment is determined to compare with the estimated torque, by assuming a virtual force $\bm{f}^{b}_{v}$ in the opposite direction of end-effector movement
\begin{equation}
    \bm{f}^{b}_{v} = -J({}^{j}\bm{q}){}^{j}\dot{\bm{q}},
\end{equation}
calculating a virtual torque using the manipulator's forward kinematics function as the moment arm,
\begin{equation}
    \bm{\tau}_{b,v} = \bm{p}^{b}_{e,j} \times \bm{f}^{b}_{v},
\end{equation}
and matching the virtual torques with the observed body torque by calculating the relative angle difference of the virtual torque and the observed torque using their dot product. The observed torque is then matched with the virtual torque with which it has the smallest direction difference. The contact point $\bm{p}_{c}$ in the world frame is then obtained by 
\begin{equation}
    \bm{p}_{c} = \bm{p}_{b} + R_{b}\bm{p}^{b}_{e,j} .
\end{equation}

\begin{figure*}[t]
    \centering
    \includegraphics[width=\linewidth]{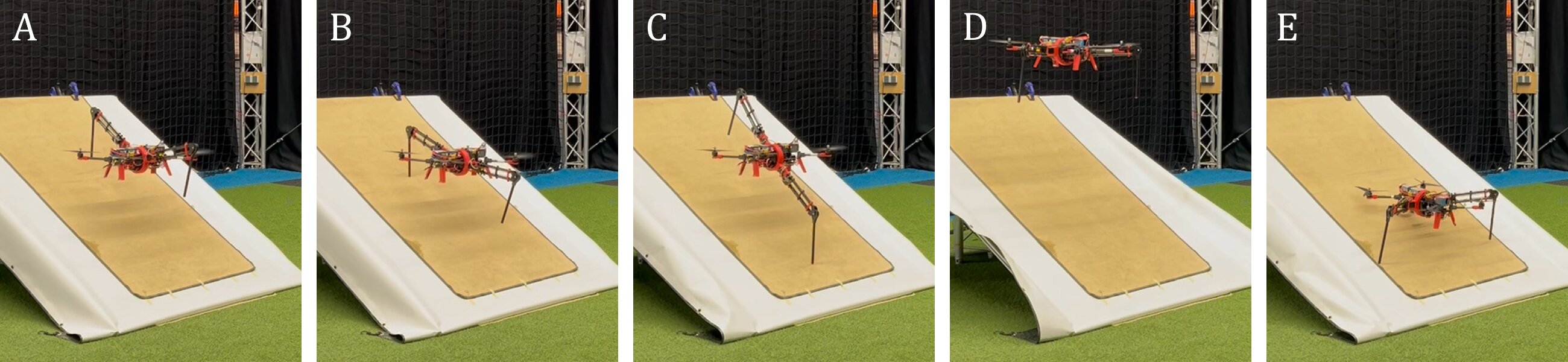}
    \caption{Image sequence showing A. Approaching the surface and setting the arms in position for probing, B. Contact on the left arm, C. Left arm raised and contact on the right arm, D. Getting in position for landing and setting the manipulator, and E. Successful landing on the inclined surface.} \label{fig:strip_figure}
\end{figure*}
\begin{figure}[t]
    \centering
    \includegraphics[width=\linewidth]{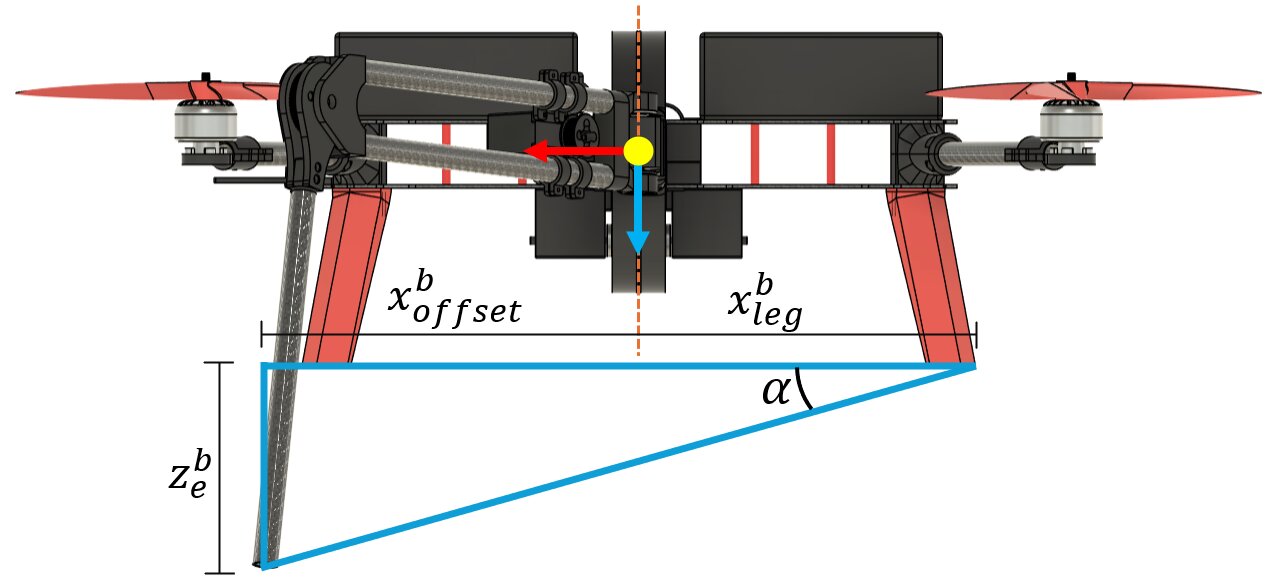}
    \caption{Side view of a CAD model of the aerial manipulator with slope-matching end-effector positions indicated for a slope of angle $\alpha$.}
    \label{fig:slope}
\end{figure}
\subsection{Surface slope determination}
For the general 3D case, a plane is fully defined by 3 points. The plane is described by a centroid, or a point on the plane (chosen to be the midpoint between the three contact points), and a surface normal, which is calculated by determining two vectors ($\bm{r}_{1},\;\bm{r}_{2}$) on the plane, taking the cross product, and normalizing to find the unit normal $\bm{n}_{s}$:
\begin{align}
\begin{aligned}
    \bm{r}_{1} = \bm{p}_{1} - \bm{p}_{2}, \\
    \bm{r}_{2} = \bm{p}_{1} - \bm{p}_{3}, \\
\end{aligned} \\
    \bm{n}_{s} = \frac{\bm{r}_{1} \times \bm{r}_{2}}{||\bm{r}_{1} \times \bm{r}_{2}||}.
\end{align}

\noindent In the experiments, only two contact points are established in a simplified case. This means that only one vector on the plane can be established. The assumption that the plane is parallel to the vehicle's x-axis, that is, the vector $\bm{e}^{b}_{1}=[1\;0\;0]^{\top}$ in the body frame is in the plane and the cross product $\bm{u}\times\bm{e}^{b}_{1}$ yields the plane's normal vector.

\subsection{Landing planning}
Once the landing plane is identified, a landing is planned. First, an approach point is selected with a certain altitude above the centroid. The required heading (i.e. surface direction) $\psi_{s}$ and surface incline $\alpha$ are determined by 
\begin{align}
    \psi_{s}=\arctan\left( \frac{n_{y}}{n_{z}} \right), \\
    \alpha = \arccos \left( \frac{||\bm{n}_{s}\cdot-\bm{e}_{3}||}{||\bm{n}_{s}||}\right),
\end{align}
where $n_{y}$ and $n_{z}$ are the y- and z-component of $\bm{n}_{s}$ and $\bm{e}_{3}=[0 \; 0 \; 1]^{\top}$, and its negative is used to find the angle w.r.t. the upwards direction.\\
\indent Then, the manipulator is moved in position such that the slope between the end-effector and the rear landing gear matches the landing surface. The end-effector position components $x^{b}_{e}$, $y^{b}_{e}$ are preset, and the required $z^{b}_{e}$ is calculated using simple trigonometry, see Figure \ref{fig:slope}. Finally, the vehicle descends to land on the slanted surface with its body in a level orientation. The full pipeline is shown in Figure \ref{fig:pipeline}.

\section{Results}
\subsection{Experimental setup}
A total of 9 experiments in 3 sets were carried out. The sets are divided by slope, with surface incline angles of $11.3^{\circ}$, $20.6^{\circ}$, and $30.5^\circ$. The angles are determined by building three slopes of differing inclination and measuring the setup when built. The slopes are covered with a high-friction material (the white and brown covering) to reduce sliding after landing. The UAM uses a motion capture  system, OptiTrack, for position feedback. \\
\indent In the experiments, the UAM is programmed to move to a start location above the surface, probe for contact in a direction (specified by a vector in the body frame), and raise the contacted arm when contact is detected to avoid finding two sequential contact points on the same arm, which would decrease accuracy of the slope angle estimation. A search pattern is implemented so that when the arms reach the end of the workspace without contact, the UAM moves slightly downwards and probes again. When two contact points are detected, the plane determination and landing planning are triggered. \\

\subsection{Experimental results}
First, one of the experiments at 30.5$^{\circ}$ incline is highlighted to provide a deeper analysis, and the latter part of this subsection provides analysis of the full set of experiments.\\
\begin{figure}[b]
    \centering
    \includegraphics[width=\linewidth]{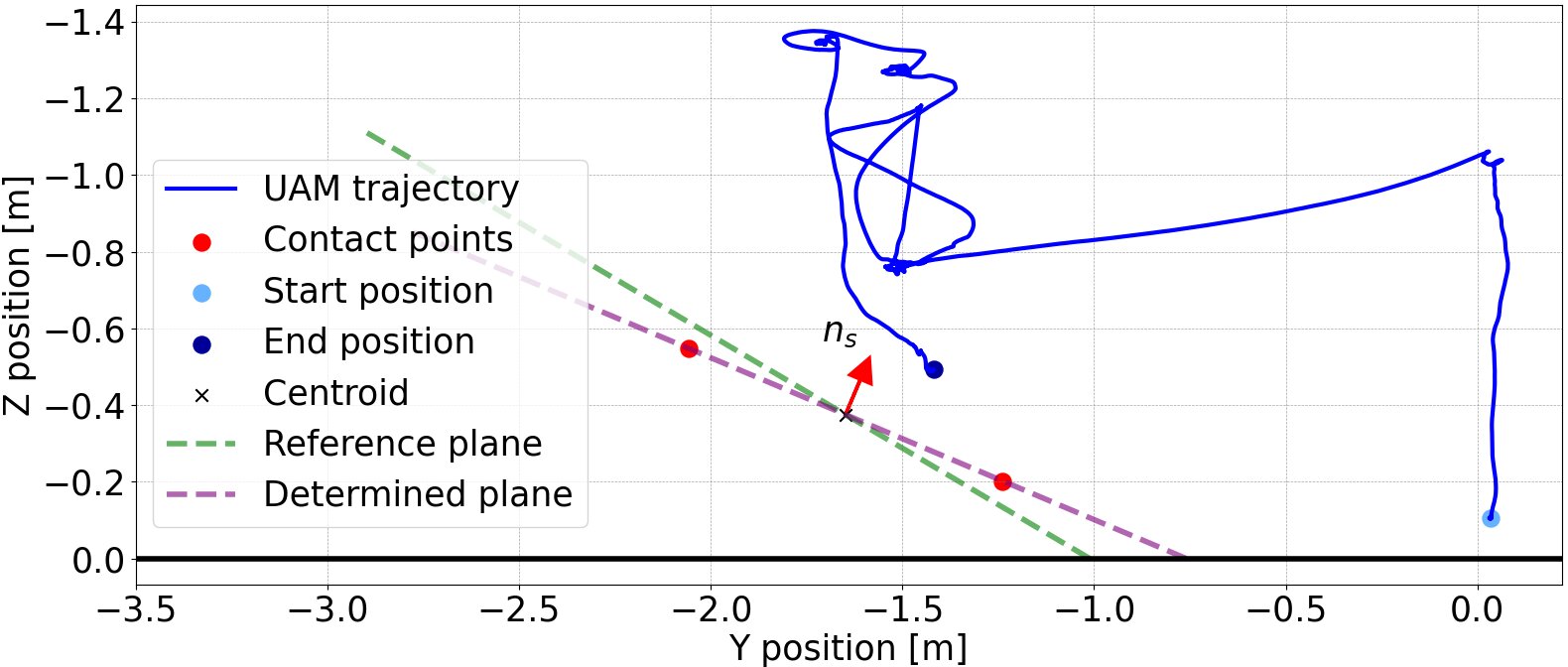}
    \caption{2D projection of the UAM position data, with the inclines of the actual plane and the plane as determined through the probing method described.}\label{fig:2dplot}
\end{figure}

\noindent Figure \ref{fig:strip_figure} provides a step-by-step view of the procedure. First, the UAM approaches the surface (Figure \ref{fig:strip_figure}.A), then it probes and finds contact on the left arm (Figure \ref{fig:strip_figure}.B). This arm is then raised to prevent multiple contact points being registered and probing continues on the right arm (Figure \ref{fig:strip_figure}.C). When both arms register a contact point, the landing is planned and the UAM moves to a location above the landing spot (Figure \ref{fig:strip_figure}.D). Finally, the arms move in position to enable the landing and the UAM lands on the rear landing gear and manipulator arms to keep a level orientation (Figure \ref{fig:strip_figure}.E).\\
\begin{figure}[t]
    \centering
    \includegraphics[width=\linewidth]{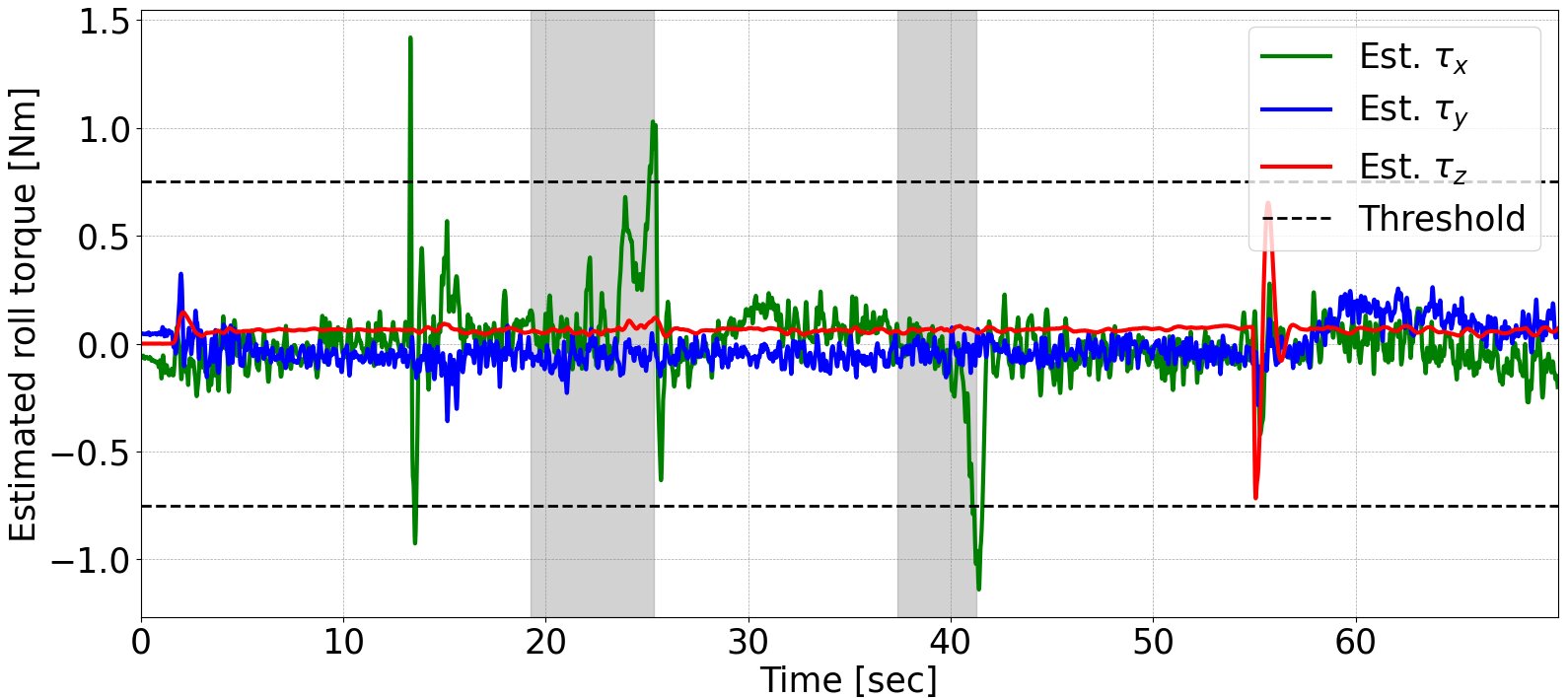}
    \caption{Estimated roll torque over time, with contact detection threshold and probing flight phases indicated in grey. The probing phase ends when the torque exceeds the threshold.}
    \label{fig:torque_time_plot}
\end{figure}
\begin{figure}[b]
    \centering
    \includegraphics[width=\linewidth]{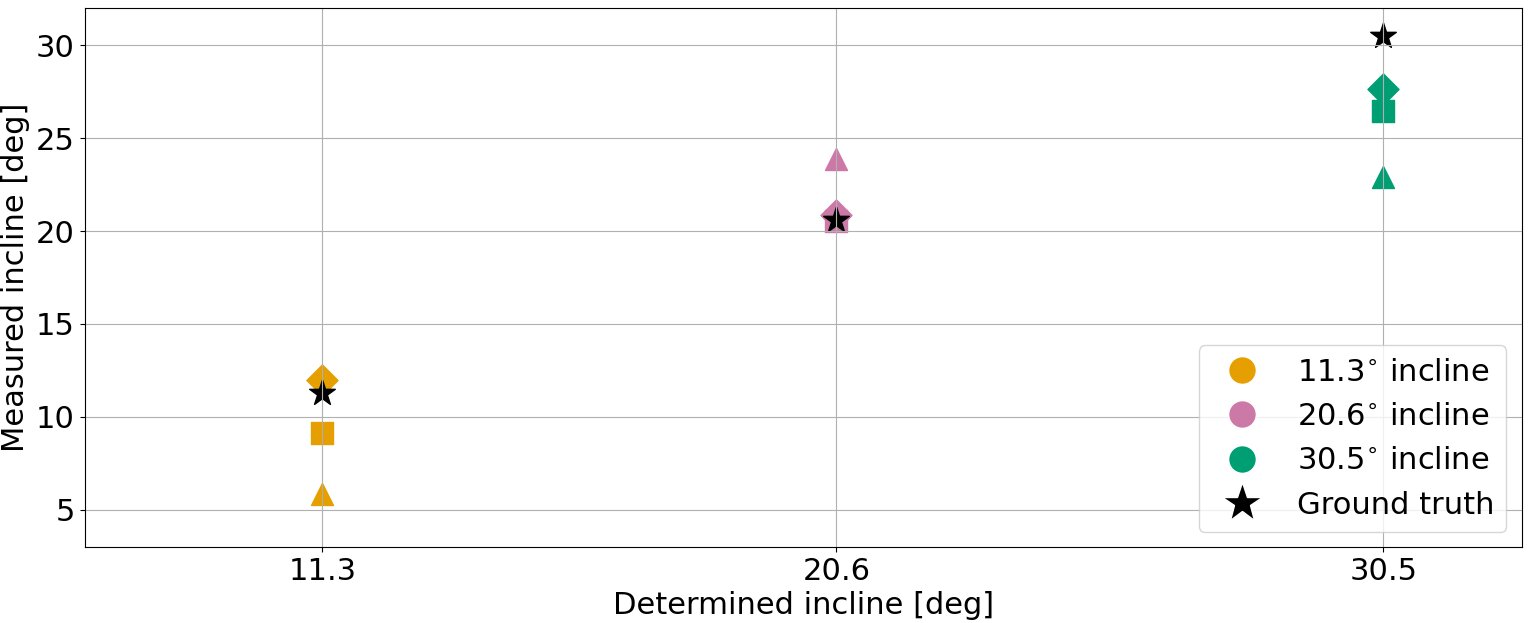}
    \caption{Estimated incline angles vs. ground truth with experiment set indicated by color, experiment number indicated by symbol, and ground truth indicated by the black stars.}
    \label{fig:scatter_all}
\end{figure}

\noindent A plot of the flight path, the surface inclines (reference and determined), and surface normal and centroid in the YZ-plane is shown in Figure \ref{fig:2dplot}. The 2D plot is chosen to highlight the difference between the actual/reference and determined plane. The UAM moves up slightly after each contact to isolate it and raise the arm that made the contact without getting stuck on the contact surface. A slight motion perpendicular to the surface happens shortly before touchdown which is attributed to the ground effect having an asymmetric effect on the vehicle due to the surface slope. It is stronger closer to the ground causing higher lift, which pitches the vehicle slightly forward causing the motion. The ground plane in the image is indicated as a bold black line.  \\
\indent The output of the torque observer is shown in Figure \ref{fig:torque_time_plot} with the fight phases indicated in shaded background. In this figure, the roll torque (in green) rises above the threshold three times before the landing sequence, of which the first time is attributed to the roll of the UAM to move from the takeoff location to the location above the surface. The registered torque here is due to inaccuracy in the system model, as in the perfect case this roll movement should not cause an external torque to be registered. This also causes the yaw torque estimation peaks near the end of the experiment as the UAM rotates to align with the plane for landing.\\

\noindent Over the 3 sets of experiments all landings were successful, meaning the UAM estimated the slope angle, set the manipulator arms in the correct configuration, and landed on the incline allowing the motors to be shut off.\\
\indent The surface incline estimations compared to the ground truth are shown in Figure \ref{fig:scatter_all}. The average estimation errors are shown in Table \ref{tab:errors}. Positive errors mean the estimate was lower than the actual value (that is, errors are calculated as real - estimated). The overall average incline estimation error is 2.87$^{\circ}$, which is sufficiently low to enable the landing robustly. \\

\noindent Comparing with the intended real-world scenario, the slope surface was intentionally covered in high-friction material, which is not usually present on roofs. This can however be easily mitigated by adding high-friction and compliant end-effectors to the manipulators, increasing independence of landing surface properties. \\

\begin{table}[]
    \centering
    \caption{Surface inclination estimation errors per set, set absolute averages and total absolute average}
    \begin{tabular}{l|l l l l}
    \hline
         & \multicolumn{4}{c}{\textbf{Error}} \\
         \textbf{Set} & \textbf{Exp. 1} & \textbf{Exp. 2} &\textbf{Exp. 3} & \textbf{Set avg. abs.} \\ \hline
         11.3$^{\circ}$ & 5.47$^{\circ}$ & 2.15$^{\circ}$ & -0.67$^{\circ}$ & 2.76$^{\circ}$ \\
         20.6$^{\circ}$ & -3.27$^{\circ}$ & 0.08$^{\circ}$ & -0.27$^{\circ}$ & 1.21$^{\circ}$ \\
         30.5$^{\circ}$ & 7.41$^{\circ}$ & 3.82$^{\circ}$ & 2.65$^{\circ}$ & 4.63$^{\circ}$  \\ \hline
         & & & \textbf{Total}  & 2.87$^{\circ}$\\ \hline
    \end{tabular}
    \label{tab:errors}
\end{table}

\section{Conclusion}
A novel omnidirectional dual-arm aerial manipulator design is introduced and demonstrated through the application of landing on slanted surfaces. To perform contact detection and determine on which of the end-effectors the contact occurs, a contact detection and localization pipeline based on a momentum-based torque estimator was developed. System validation is done through a flight testing campaign comprising of 9 total flight tests on 3 different surface inclines, showing effectiveness up to 30.5$^{\circ}$ slope angle. The average error of the surface inclination estimation algorithm is $2.87^{\circ}$, which was shown to be sufficient for repeatably planning and executing landings. \\
\indent The assumption that the plane is aligned with the UAM's heading will be addressed in future work through adding contact points, constraining the plane in 3D space. In addition, advanced application scenarios will be explored to demonstrate the versatility of the aerial manipulator’s novel morphology, particularly in use cases that require coordinated dual-arm manipulation and navigation.

\section*{Acknowledgement}
The authors would like to thank Anton Bredenbeck and Dimitris Chaikalis for their help with the hardware and software.

\addtolength{\textheight}{-8.5cm}   

\newpage
\bibliographystyle{IEEEtran}
\bibliography{IEEEabrv,references.bib}
\end{document}